\documentclass[pdflatex,sn-mathphys-num]{sn-jnl}% Math and Physical Sciences Numbered Reference Style
\usepackage{graphicx}%
\usepackage{multirow}%
\usepackage{amsmath,amssymb,amsfonts}%
\usepackage{amsthm}%
\usepackage{mathrsfs}%
\usepackage[title]{appendix}%
\usepackage{xcolor}%
\usepackage{textcomp}%
\usepackage{manyfoot}%
\usepackage{booktabs}%
\usepackage{algorithm}%
\usepackage{algorithmicx}%
\usepackage{algpseudocode}%
\usepackage{listings}%

\usepackage[acronym,nonumberlist]{glossaries}
\newacronym{AD}{AD}{Autonomous Driving}
\newacronym{SLR}{SLR}{Systematic Literature Review}
\newacronym{CP}{CP}{Collaborative Perception}
\newacronym{CD}{CD}{Cooperative Driving}
\newacronym{ETSI}{ETSI}{European Telecommunications Standards Institute}
\newacronym{PDU}{PDU}{Packet Data Unit}
\newacronym{ITS}{ITS}{Intelligent Transportation System}
\newacronym{C-ITS}{C-ITS}{Cooperative-Intelligent Transportation Systems}
\newacronym{CPM}{CPM}{Cooperative Perception Message}
\newacronym{MCM}{MCM}{Maneuver Coordination Message}
\newacronym{DSRC}{DSRC}{Dedicated Short-Range Communication}
\newacronym{C-V2X}{C-V2X}{Cellular Vehicle-to-Everything}
\newacronym{LSTM}{LSTM}{Long Short-Term Memory}
\newacronym{V2X}{V2X}{Vehicle-to-Everything}
\newacronym{V2V}{V2V}{Vehicle-to-Vehicle}
\newacronym{V2I}{V2I}{Vehicle-to-Infrastructure}
\newacronym{IoV}{IoV}{Internet of Vehicles}
\newacronym{VRUs}{VRUs}{Vulnerable Road Users}
\newacronym{FOV}{FOV}{Field of View}
\newacronym{ROIs}{ROIs}{regions of interest}

\newacronym{AVs}{AVs}{Autonomous Vehicles}
\newacronym{CAVs}{CAVs}{Connected Autonomous Vehicles}
\newacronym{CAV}{CAV}{Connected Autonomous Vehicle}
\newacronym{BEV}{BEV}{Bird’s Eye View}
\newacronym{GNN}{GNN}{Graph Neural Network}

\newacronym{COD}{COD}{Collaborative Object Detection}
\newacronym{COT}{COT}{Collaborative Object Tracking}
\newacronym{CMP}{CMP}{Collaborative Motion Prediction}
\newacronym{CSS}{CSS}{Collaborative Semantic Segmentation}
\newacronym{CLD}{CLD}{Collaborative Lane Detection}

\newacronym{PandP}{P\&P}{Joint Perception and Prediction}
\newacronym{Co-PandP}{Co-P\&P}{Collaborative Joint Perception and Prediction}

\newacronym{HD Map}{HD Map}{High Definition Map}
\newacronym{GNSS}{GNSS}{Global Navigation Satellite System}
\newacronym{GPS}{GPS}{Global Positioning System}
\newacronym{SLAM}{SLAM}{Simultaneous Localization and Mapping}
\newacronym{CSC}{CSC}{Collaborative Scene Completion}
\newacronym{RV}{RV}{Range-View}
\newacronym{IoU}{IoU}{Intersection-over-Union}

\theoremstyle{thmstyleone}%
\theoremstyle{thmstyletwo}%

\theoremstyle{thmstylethree}%

\begin{document}

\title[Towards Collaborative Joint Perception and Prediction: Framework, Baseline Evaluation, and Deployment Perspectives]{Towards Collaborative Joint Perception and Prediction: Framework, Baseline Evaluation, and Deployment Perspectives}

\ifdefined\iflatexml\else\newif\iflatexml\latexmlfalse\fi

\iflatexml

  \providecommand{\fnm}[1]{#1}
  \providecommand{\sur}[1]{#1}
  \providecommand{\email}[1]{}
  \author{%
    Lei Wan$^{1,2}$\thanks{Corresponding author: lei.wan@xitaso.com} \and
    Hannan Ejaz Keen$^{1}$ \and
    Alexey Vinel$^{2}$\\[4pt]
    \small $^{1}$XITASO GmbH, Augsburg, 86153, Germany\\
    \small $^{2}$Karlsruhe Institute of Technology, Karlsruhe, 76131, Germany}
\else

  \author*[1,2]{\fnm{Lei} \sur{Wan}}\email{lei.wan@xitaso.com}
  \author[1]{\fnm{Hannan Ejaz} \sur{Keen}}\email{hannan.keen@xitaso.com}
  \author[2]{\fnm{Alexey} \sur{Vinel}}\email{alexey.vinel@kit.edu}
  \affil[1]{\orgname{XITASO GmbH}, \city{Augsburg}, \postcode{86153}, \country{Germany}}
  \affil[2]{\orgname{Karlsruhe Institute of Technology}, \city{Karlsruhe}, \postcode{76131}, \country{Germany}}
\fi

%%=============================================================%%
%% GivenName	-> \fnm{Joergen W.}
%% Particle	-> \spfx{van der} -> surname prefix
%% FamilyName	-> \sur{Ploeg}
%% Suffix	-> \sfx{IV}
%% \author*[1,2]{\fnm{Joergen W.} \spfx{van der} \sur{Ploeg} 
%%  \sfx{IV}}\email{iauthor@gmail.com}
%%=============================================================%%

%% \author*[1,2]{Lei Wan}
%% \email{lei.wan@xitaso.com}

%% \author[1]{Hannan Ejaz Keen}
%% \email{hannan.keen@xitaso.com}

%% \author[2]{Alexey Vinel}
%% \email{alexey.vinel@kit.edu}

%% \affil*[1]{XITASO GmbH, Augsburg, 86153, Germany}

%% \affil[2]{Karlsruhe Institute of Technology, Karlsruhe, 76131, Germany}

%% \author*[1,2]{\fnm{Lei} \sur{Wan}}\email{lei.wan@xitaso.com}

%% \author[1]{\fnm{Hannan Ejaz} \sur{Keen}}\email{hannan.keen@xitaso.com}

%% \author[2]{\fnm{Alexey} \sur{Vinel}}\email{alexey.vinel@kit.edu}

%% \affil[1]{\orgname{XITASO GmbH}, \orgaddress{\city{Augsburg}, \postcode{86153}, \country{Germany}}}

%% \affil*[2]{\orgname{Karlsruhe Institute of Technology}, \orgaddress{\city{Karlsruhe}, \postcode{76131}, \country{Germany}}}

%%==================================%%
%% Sample for unstructured abstract %%
%%==================================%%

\abstract{\acrfull{CAVs} increasingly exploit \acrfull{V2X} communication to exchange multi-source sensor information, enabling advanced \acrfull{CP} capabilities. Extending beyond these capabilities, this work focuses on \acrfull{Co-PandP}, a paradigm that unifies \acrshort{CP} with motion prediction to mitigate two persistent challenges: the accumulation of perception errors and visual occlusions. We present a conceptual framework for \acrfull{Co-PandP} that improves motion prediction of surrounding road users, thereby enhancing situational awareness in complex and dynamic traffic environments \cite{vehits25}. Building upon our preliminary study, this extended version compares the performance of different fusion strategies and establishes baseline performance for a modular design of perception and prediction. Experimental results show that prediction-level fusion leads to a decline in overall system performance compared to detection-level or tracking-level fusion. We further implement a minimal end-to-end Co-P\&P prototype that couples collaborative point-cloud sharing via the RENO neural codec with joint detection--forecasting via FutureDet, showing that collaboration improves forecasting accuracy while neural compression preserves this benefit at roughly $34\times$ lower communication bandwidth.}


%%\pacs[JEL Classification]{D8, H51}

%%\pacs[MSC Classification]{35A01, 65L10, 65L12, 65L20, 65L70}

\maketitle

\section{Introduction} \label{sec:introduction}

\begin{table}[t]
\centering
\small
\caption{Nomenclature: abbreviations used in this paper.}
\label{tab:abbrev}
\begin{tabular}{ll}
\toprule
Abbreviation & Definition \\
\midrule
AD      & Autonomous Driving \\
AVs     & Autonomous Vehicles \\
BEV     & Bird's Eye View \\
CAVs    & Connected Autonomous Vehicles \\
C-ITS   & Cooperative Intelligent Transportation System \\
Co-P\&P & Collaborative Joint Perception and Prediction \\
CP      & Collaborative Perception \\
CPM     & Cooperative Perception Message \\
CSC     & Collaborative Scene Completion \\
C-V2X   & Cellular Vehicle-to-Everything \\
DSRC    & Dedicated Short-Range Communication \\
FOV     & Field of View \\
GNSS    & Global Navigation Satellite System \\
GPS     & Global Positioning System \\
HD Map  & High Definition Map \\
IoU     & Intersection-over-Union \\
LSTM    & Long Short-Term Memory \\
P\&P    & Joint Perception and Prediction \\
ROIs    & Regions of Interest \\
RV      & Range-View \\
V2I     & Vehicle-to-Infrastructure \\
V2V     & Vehicle-to-Vehicle \\
V2X     & Vehicle-to-Everything \\
\bottomrule
\end{tabular}
\end{table}

\gls{AD} technology plays a pivotal role in the evolution of intelligent transportation systems, offering the potential to enhance road safety, improve traffic flow, conserve energy, and reduce carbon emissions. At the core of an \gls{AD} system lies the perception module, responsible for detecting dynamic objects and understanding the static surroundings. This module integrates multiple tasks, such as object detection, tracking, motion prediction, and semantic segmentation, which have been extensively studied in prior research supported by public datasets \cite{geiger2012we,caesar2020nuscenes,sun2020scalability,mirlach2025r,keen2021drive}. Traditionally, these tasks have been organized in a modular pipeline, forming the perception foundation for downstream operations such as motion planning and control \cite{10.1007/978-3-031-32606-6_23,10.1007/978-3-030-19648-6_56}. Although advances in artificial intelligence and multi-sensor fusion have considerably improved perception capabilities, single-vehicle perception remains vulnerable to limitations, visual occlusions, which can compromise safety and lead to hazardous situations.

Beyond onboard perception, connected vehicles operate within a broader intelligent-transportation context in which network-level information from roadside and traffic sensors supports routing and traffic management; for example, sensor-augmented routing in stochastic networks~\cite{owais_pareto, almutairi_routing} and network-level delay estimation from passive traffic sensors~\cite{owais_delay} illustrate how infrastructure-derived information complements onboard vehicle sensing.

\gls{V2X} communication provides an effective means to overcome such shortcomings by enabling vehicles to share sensory information with other vehicles and infrastructure, thereby extending perception beyond the line of sight. Through \gls{V2X}, \gls{CAVs} can perform \gls{CP}, leveraging data from multiple viewpoints to achieve a more holistic understanding of the environment. Early studies in \gls{CP} originated from the communications domain, emphasizing message standardization and transmission efficiency. More recently, the focus has shifted toward computer vision and robotics, where research is moving beyond the exchange of standardized perception messages, such as \gls{CPM} containing detected object lists, toward the sharing of raw sensory data or intermediate neural features. For example, Chen et al. \cite{FcooperFeatureBased-2019-chend} proposed a LiDAR feature-sharing framework that fuses point cloud features from multiple vehicles, achieving performance gains under bandwidth constraints. Likewise, Hu et al. \cite{CollaborationHelpsCamera-2023-hub} demonstrated a camera-based \gls{CP} approach that aggregates multi-agent \gls{BEV} visual features to provide richer scene understanding of dynamic road users.

\begin{figure}[t]
\centering
\includegraphics[width=0.5\linewidth]{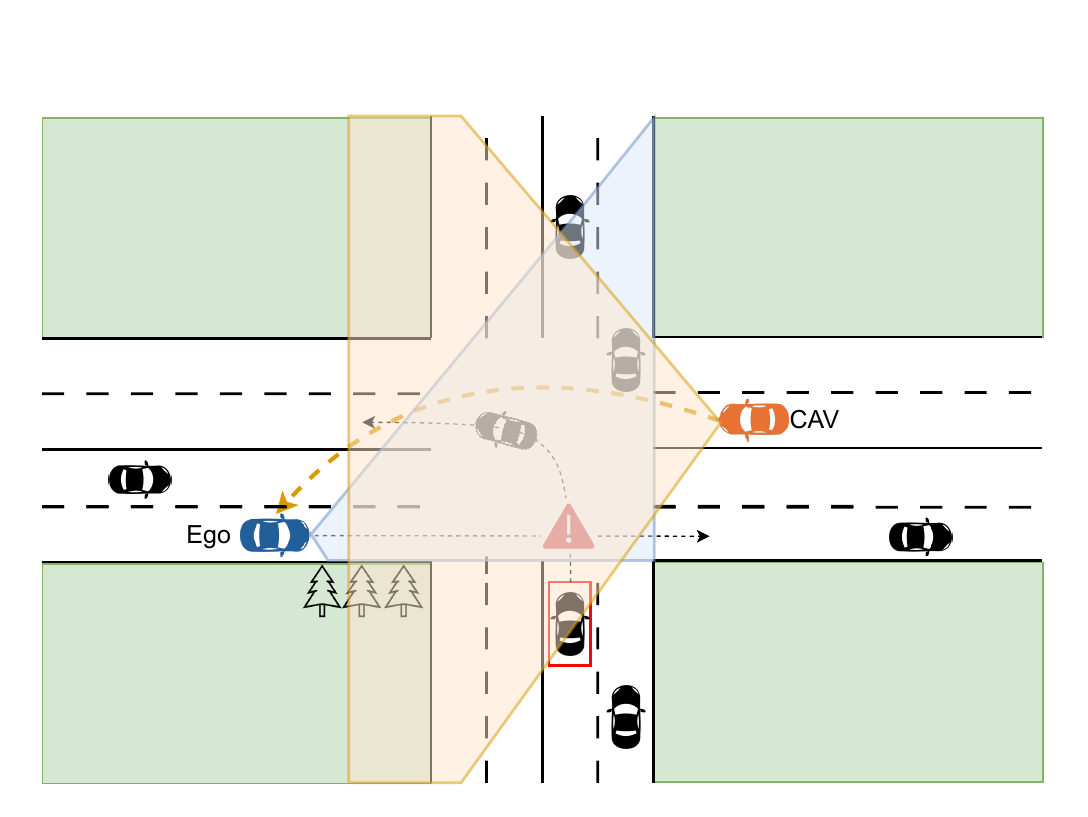}
\caption{Schematic diagram of Collaborative Perception (adapted from \cite{vehits25}). The illustration depicts an intersection scenario where two \gls{CAVs} cooperate to improve perception. The ego vehicle (blue) faces occlusions from trees and buildings, obscuring a left-turning vehicle. A collaborating vehicle (orange) located across the intersection shares its sensor data, extending the ego vehicle’s awareness. Shaded areas indicate each vehicle’s \gls{FOV}.}
\label{fig:cp}
\end{figure}

The scope of \gls{CP} extends well beyond object detection, encompassing a range of other perception tasks. For example, Liu et al. \cite{LiDARSemanticSegmentation-2023-liua} propose a collaborative semantic segmentation framework that leverages intermediate feature sharing to achieve performance surpassing single-vehicle baselines. In the domain of motion prediction, Wang et al. \cite{V2VNetVehicletoVehicleCommunication-2020-wangb} demonstrate that inter-vehicle collaboration can substantially improve trajectory prediction accuracy. Certain perception tasks are also addressed within multi-task learning frameworks, for instance, the V2XFormer introduced by Wang et al. \cite{DeepAccidentMotionAccident-2024-wanga} produces detection, motion prediction, and semantic segmentation outputs simultaneously. While multi-task designs benefit from computational efficiency by sharing a common feature extractor, they often fail to incorporate temporal dependencies across sensor frames, an aspect that is crucial for reliable tracking and accurate motion forecasting.

A notable research direction involves the creation of differentiable frameworks that unify multiple perception tasks into a single, end-to-end trainable model. Liang et al. \cite{liang2020pnpnet}, for instance, present an end-to-end \gls{PandP} framework tailored for LiDAR-equipped vehicles, while Gu \cite{gu2023vip3d} develops a vision-only end-to-end \gls{PandP} pipeline that uses visual features to jointly perform detection and motion prediction. These end-to-end paradigms offer a promising means to overcome the limitations of traditional modular perception pipelines, where error accumulation across sequential stages can degrade overall performance. By learning tasks jointly, such frameworks help suppress cumulative noise and allow motion prediction to benefit from richer, fine-grained contextual cues. Nevertheless, existing \gls{PandP} research continues to face notable hurdles, with visual occlusion remaining a critical factor that can significantly reduce prediction reliability for targets hidden from direct view.

To address these limitations, this paper proposes the \gls{Co-PandP} framework, leveraging \gls{V2X}-enabled collaboration \cite{vehits25}. The underlying principle is that collaborative perception (\gls{CP}) naturally complements the perception capability of individual vehicles, enabling the framework to operate flexibly in both \gls{V2X}-enabled and standalone scenarios. Furthermore, the proposed approach separates the collaborative scene representation from subsequent perception tasks to facilitate easier deployment and to improve scalability. Drawing inspiration from recent collaborative reconstruction frameworks \cite{MultiRobotSceneCompletion--lib,CoreCooperativeReconstruction-2023-wanga}, our method uses collaborative scene completion (\acrfull{CSC}) to effectively address occlusion-related perception issues. Thus, the resulting \gls{Co-PandP} framework is structured around two independent but complementary modules: the \acrfull{CSC} module and the \acrfull{PandP} module.

% In parallel to the development of new \gls{CP} methods, there is also a demand for improved evaluation frameworks that can accurately measure collaborative performance. Current evaluation methodologies primarily originate from single-agent scenarios and do not fully capture the advantages provided by collaborative approaches. Novel evaluation metrics tailored explicitly for the proposed \gls{Co-PandP} framework are essential to fairly assess its performance and effectiveness.

The primary contributions of this work are summarized as follows:
\begin{itemize}
\item A novel conceptual framework, \gls{Co-PandP}, designed specifically to mitigate cumulative errors in modular perception pipelines and overcome challenges due to visual occlusion.
\item An analysis on optimal timing for performing \gls{V2X} fusion by comparing outcomes obtained from fusion at different stages of the perception pipeline.
\item The implementation and benchmarking of a minimal end-to-end \gls{Co-PandP} prototype, coupling RENO-based~\cite{you2025reno} collaborative point-cloud sharing with joint detection--forecasting via FutureDet~\cite{peri2022forecasting}, which provides a first empirical instantiation of the framework beyond the modular pipeline (quantitative results in Section~\ref{sec:result}).
\item An in-depth discussion on limitations of existing evaluation methods for collaborative perception and the introduction of a new evaluation methodology tailored explicitly to \gls{Co-PandP}, emphasizing the value of \gls{V2X}-based collaboration.
\item A detailed exploration of open challenges in the deployment and practical implementation of the \gls{Co-PandP} framework, along with recommendations for future research directions.
\end{itemize}

The remainder of this paper is structured as follows: Section~\ref{sec:related_work} reviews related literature. Section~\ref{sec:system} provides a detailed description of our proposed system design. In Section~\ref{sec:evaluation}, we introduce our evaluation methodology tailored specifically to the \gls{Co-PandP} framework. Section~\ref{sec:result} reports experimental results along with their analysis. Section~\ref{sec:challenges} identifies practical deployment challenges encountered in realistic settings. Finally, Section~\ref{sec:conclusion} summarizes our findings and outlines directions for future research.

\textbf{Relation to our prior work.}
This article is an extended version of our earlier conference
paper~\cite{vehits25}, which introduced the conceptual Co-P\&P
architecture and its constituent modules. The present manuscript substantially extends
\cite{vehits25} along five directions: (i) we provide a concrete modular
instantiation of the perception-and-prediction pipeline and report the
first quantitative baseline on a real-world cooperative dataset; (ii) we
implement and benchmark a minimal \emph{end-to-end} Co-P\&P prototype, in
which collaborative point-cloud sharing is realized with the RENO~\cite{you2025reno}
neural codec and joint detection--forecasting with FutureDet~\cite{peri2022forecasting},
and show that collaboration improves forecasting accuracy while neural
compression preserves this benefit at up to $34\times$ lower per-frame
bandwidth; (iii) we conduct a systematic comparison of V2X fusion performed
at the detection, tracking, and prediction stages, including a joint
accuracy/communication-cost analysis, and show that prediction-level fusion
is detrimental; (iv) we operationalize a visibility-stratified evaluation
tailored to collaborative perception; and (v) we add an in-depth analysis of
deployment challenges and a dedicated Discussion that contextualizes the
findings. The conceptual framework is summarized here only for
self-containedness, while all experimental results, the end-to-end
prototype, the fusion-timing study, the comparison with prior frameworks,
and the Discussion are new to this article.

% introduction of CAVs and V2X
% introduction of CP
% introduce the COD,COT, CMP
% introduction of joint P&P
% introduction of this paper

\section{Related Work} \label{sec:related_work}
This section reviews the literature on \gls{PandP}, \gls{Co-PandP}, and \gls{CSC}. We begin by discussing single-agent \gls{PandP} approaches that unify detection, tracking, and prediction tasks within a single framework. Then, we explore recent advancements in \gls{Co-PandP}, which address the limitations of individual agents through V2X collaboration. Finally, we summarize methods in \gls{CSC}, which provide upstream task-agnostic scene representations for supporting various downstream tasks in collaborative settings.
\subsection{\acrfull{PandP}}
\gls{PandP} demonstrate significant potential in addressing the cumulative errors and high computational demands characteristic of modular perception pipelines by unifying multiple tasks within the framework \cite{dal2024joint}. For instance, Luo \cite{luo2018fast} introduces the first fully end-to-end neural network that simultaneously handles 3D object detection, tracking, and trajectory prediction based on \gls{BEV} LiDAR representations. Building upon this, Liang et al. \cite{liang2020pnpnet} propose PnPNet, explicitly integrating tracking with detection and prediction through \gls{LSTM} networks to effectively model temporal dependencies, greatly enhancing prediction accuracy. Peri et al. \cite{peri2022forecasting} present FutureDet, designed to forecast object states in future, unobserved LiDAR frames, linking these predictions to current detections and enabling inherently multimodal forecasts, where the model accounts for multiple plausible future trajectories due to inherent uncertainties in dynamic environments. Beyond \gls{BEV}, LiDAR data can also be represented using panoramic \gls{RV} images, which encode LiDAR measurements in a cylindrical projection, capturing the full 360-degree horizontal field of view. LaserFlow \cite{meyer2020laserflow} introduces the first \gls{RV}-based \gls{PandP} approach, utilizing a multi-sweep fusion strategy with 2D convolutions to provide probabilistic trajectory predictions. Another representation approach involves 3D voxel grids, which segment the environment into uniform cubic cells, preserving detailed spatial geometries. Liu et al. \cite{liu2024lidar} present Occ4cast, the first occupancy-grid-based \gls{PandP} approach, employing 3D convolutions and \gls{LSTM} architectures to model spatio-temporal dynamics and utilizing self-supervised learning for improved predictive outcomes.

Camera-based methods have also become increasingly prominent for \gls{PandP}. Zhang et al. \cite{zhang2022beverse} introduce BEVerse, using multi-view camera images to construct semantic maps, employing cell-wise latent vectors for capturing uncertainties. Similarly, Gu et al. \cite{gu2023vip3d} present ViP3D, integrating agent-centric 3D queries, graph neural networks, and vectorized \gls{HD Map} representations for multimodal trajectory forecasting. Within end-to-end driving frameworks, unified neural network solutions for \gls{PandP} have emerged; for instance, Hu et al. \cite{hu2023planning} introduce UniAD, which integrates waypoint prediction and occupancy forecasting through semantic map queries and the transformer-based OccFormer architecture, explicitly capturing interactions among road agents.

While LiDAR-based approaches excel at precisely estimating 3D object locations and shapes, they lack contextual richness due to the sparse nature of point clouds. Conversely, camera-based methods provide rich visual context but typically struggle with precise 3D localization due to limited depth information. To leverage the advantages of both modalities, multi-representation fusion methods have gained attention. Khalil et al. \cite{9738812} introduce LiCaNet, a fusion architecture combining BEV LiDAR data with front- and rear-view images, achieving a more robust and comprehensive scene understanding. LiCaNet utilizes the Spatio-Temporal Pyramid Network (STPN) from MotionNet \cite{wu2020motionnet} for feature extraction, leveraging its triple-output architecture, comprising cell classification, motion prediction, and state estimation tasks, as foundational components. Following a similar fusion paradigm within an end-to-end driving framework, Ye et al. \cite{ye2023fusionad} present FusionAD, an enhancement of UniAD, merging LiDAR and multi-view images through advanced transformer-based attention mechanisms, thereby achieving superior performance in multimodal trajectory forecasting and occupancy prediction tasks.

\subsection{\acrfull{Co-PandP}}
Research on \gls{Co-PandP} remains in its early stages, with only a limited number of studies exploring this direction. The concept aims to overcome key limitations of conventional \gls{PandP} frameworks, such as visual occlusions and restricted long-range perception, by leveraging multi-agent collaboration. Zhou et al. \cite{zhou2024v2xpnp} propose an intermediate fusion framework employing a single-step communication strategy for \gls{Co-PandP}, combining multi-frame and multi-agent data through spatio-temporal fusion to improve performance in V2X-enabled scenarios. Additionally, Yu et al. \cite{yu2025end} introduce UniV2X, an end-to-end collaborative driving model based on UniAD \cite{hu2023planning}, leveraging multi-task learning to simultaneously address both perception and prediction tasks. Among these, V2XPnP~\cite{zhou2024v2xpnp} is the framework most directly comparable to our collaborative joint perception-and-prediction setting. However, it currently provides no public code release and reports no benchmark results on DAIR-V2X-Seq, so a fair head-to-head quantitative comparison is not yet feasible; we therefore position our study as a controlled, reproducible baseline and leave such a comparison to future work once a public implementation or shared benchmark becomes available.

\subsection{\acrfull{CSC}}
In recent years, a few pioneering works have explored the use of scene completion for \gls{V2X} collaboration. \gls{CSC} aims to provide a task-agnostic foundation for \gls{CP} by developing upstream modules that can support a wide range of downstream perception tasks \cite{wan2025systematic}. For example, Li \cite{MultiRobotSceneCompletion--lib} proposes the spatio-temporal autoencoder (STAR), a multi-robot framework designed for collaborative scene completion. Similarly, Wang et al. \cite{wang2023core} introduce CORE, a cooperative reconstruction approach where perception modules leverage a shared backbone whose parameters are jointly learned during the cooperative reconstruction training phase.

\section{Details of the System} \label{sec:system}
Figure~\ref{fig:cp_framework} illustrates the conceptual architecture of the proposed framework, which integrates sensing, localization, \gls{HD Map}, communication, \gls{PandP}, and collaborative scene completion (\gls{CSC}). The framework combines GPS/GNSS-INS and map-based localization to achieve accurate positioning. Sensor data, including LiDAR point clouds and poses, from both collaborating vehicles and roadside units are transmitted to the ego vehicle. A \gls{CSC} module aggregates these data sources to generate a unified, comprehensive LiDAR frame. To optimize bandwidth efficiency, intermediate feature representations are exchanged instead of raw point clouds via \gls{V2X} communication. A collaboration trigger controls the activation of \gls{CSC}. The \gls{PandP} module then fuses localization, map, and LiDAR information to jointly perform perception and prediction, providing results that feed into risk assessment as well as planning and control modules for real-time decision-making. This section provides an
overview of each core component of the framework along with their corresponding
approaches.

\begin{figure*}[ht]
\centering
\includegraphics[width=\linewidth]{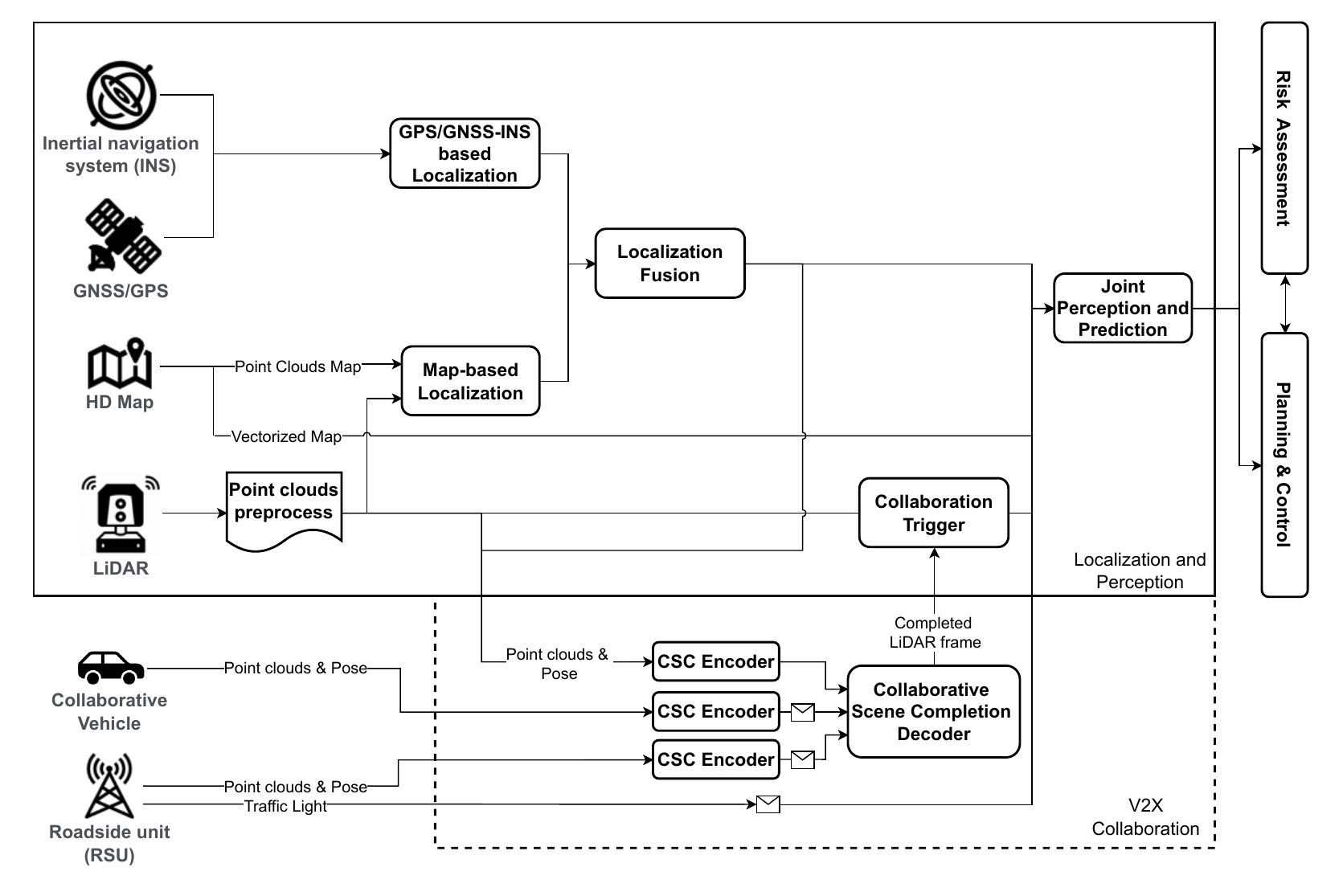}
\caption{Conceptual architecture of the proposed \gls{Co-PandP} system. Sensor data and poses from the ego vehicle, collaborating vehicles, and roadside units feed a CSC module and a \gls{PandP} module, whose outputs drive risk assessment and planning.}
\label{fig:cp_framework}
\end{figure*}

\subsection{Sensing}

A variety of sensing modalities can be employed to obtain a 3D representation of the environment, including LiDAR, radar, and cameras such as RGB and infrared. In the proposed framework, LiDAR is selected as the primary sensing modality because of its high accuracy in 3D range measurements, which greatly enhances the spatial perception capabilities of \gls{AVs}.

LiDAR sensors differ in their scanning mechanisms, commonly categorized into spinning and oscillating types \cite{triess2021survey}. Spinning LiDAR employs a uniform rotational scanning pattern that evenly distributes points across a $360^\circ$ \gls{FOV}. In contrast, oscillating LiDAR follows a snake-like scanning trajectory, producing denser yet less uniform point distributions within a limited \gls{FOV}. These variations result in distinct point cloud characteristics, which can introduce a domain gap for perception models when transferring between sensor types. Effectively addressing such domain gaps is essential for developing robust \gls{CP} systems that operate reliably across heterogeneous LiDAR configurations.

\subsection{Localization}

In addition to environmental perception, precise self-localization is a critical prerequisite for \gls{CP}. Accurate positioning enables data from multiple dynamic agents to be fused within a shared coordinate system, ensuring proper alignment of all sensory inputs. Consequently, the overall effectiveness of \gls{CP} is strongly dependent on the localization accuracy of participating \gls{CAVs}.

Conventional vehicle localization often relies on \gls{GNSS} or \gls{GPS} to estimate position through trilateration. However, \gls{GNSS}-based methods are susceptible to issues such as Non-Line-of-Sight and multipath propagation, which can produce errors exceeding 3 meters, significantly degrading the reliability and safety of \gls{AD} systems \cite{OCHIENG2002171}. \gls{HD Map} technology can help mitigate such errors, achieving centimeter-level positioning accuracy \cite{chalvatzaras2022survey}. These maps are typically generated via extensive mapping campaigns, frequently employing LiDAR to create high-resolution point cloud layers. In practice, precise vehicle localization on an \gls{HD Map} is achieved by integrating \gls{GNSS} measurements with LiDAR-based localization, yielding a robust and accurate positioning solution.

\subsection{\gls{HD Map}}

An \gls{HD Map} serves not only as a basis for localization but also as a rich source of semantic information describing the static environment. It contains detailed road attributes such as lane boundaries, centerlines, road markings, traffic signs, poles, and traffic light positions. This semantic context supports vehicles in interpreting traffic rules and understanding their surroundings, thereby improving motion prediction accuracy. Xu et al. \cite{Xu2023TowardsMF} demonstrate the strong influence of map quality on prediction performance, showing that high-quality, curated \gls{HD Map} datasets consistently outperform systems relying on online-generated maps or operating without maps.  

In the proposed framework, the \gls{HD Map} operates as an independent module interfacing with the perception subsystem. This modular design ensures compatibility with diverse mapping solutions and facilitates scalability, enabling integration with online mapping pipelines or even deployment in cost-sensitive scenarios using lightweight, mapless approaches.

\subsection{Communication}

\gls{V2X} communication serves as a fundamental enabler for \gls{CP}, allowing \gls{CAVs} and intelligent infrastructure to share sensor-derived environmental information. Two primary technologies currently support \gls{V2X} communications: \gls{DSRC} and cellular-based solutions \cite{abboud2016interworking}.

\gls{DSRC} is a wireless standard developed for automotive and \gls{C-ITS} applications, enabling short-range, low-latency data exchange without the need for external network infrastructure. Its minimal latency makes it well-suited for safety-critical applications \cite{5888501}. However, \gls{DSRC} is constrained by its limited communication range and reduced scalability in dense traffic environments \cite{harding2014vehicle}. \gls{C-V2X} offers a complementary alternative, providing higher bandwidth and extended coverage. These features allow transmission of large volumes of sensor data essential for \gls{CP}. Nevertheless, certain \gls{C-V2X} modes rely on cellular infrastructure, which can lead to increased latency or degraded performance in areas far from base stations.

Given the respective limitations of \gls{DSRC} and \gls{C-V2X} when used independently, hybrid communication strategies combining both technologies are increasingly attractive. Such \gls{DSRC}–cellular interworking schemes can leverage the strengths of each: low-bandwidth information such as traffic light status can be efficiently transmitted over \gls{DSRC}, while high-bandwidth sensor data is better suited for \gls{C-V2X}.

\subsection{Joint Perception and Prediction}

The \gls{PandP} module forms the central component of the proposed framework. It integrates multi-source data, including LiDAR point clouds, vehicle pose, \gls{HD Map} information, and traffic light status, to jointly perform object detection, tracking and motion forecasting for surrounding agents, as illustrated in Figure~\ref{fig:pandp_framework}. The \gls{PandP} processing pipeline comprises a LiDAR encoder, temporal encoder, map encoder, multi-agent interaction attention module, and a \gls{PandP} decoder. The module takes as input completed LiDAR frames from the \gls{CSC} module or raw LiDAR frames from ego, vehicle pose, \gls{HD Map}, and traffic light information. A collaboration trigger determines when collaborative perception is activated. LiDAR data are processed through a LiDAR encoder and spatial-temporal attention to extract dynamic scene features, while map and traffic light information are processed through a dedicated map encoder. These feature streams are fused in the multi-agent interaction attention block to model both agent–map and inter-agent relationships. The \gls{PandP} decoder integrates all features to jointly output perception results (object detection and tracking) and motion forecasts for surrounding agents.

\begin{itemize}
    \item \textbf{LiDAR Encoder}: The LiDAR encoder extracts semantic features from raw point clouds to facilitate scene understanding for \gls{AVs}. For instance, VoxelNet~\cite{zhou2018voxelnet} partitions the point cloud into a 3D voxel grid, aggregates features within each voxel, and applies voxel-wise convolutions to capture local 3D spatial structures. These features are subsequently collapsed along the vertical axis and converted into \gls{BEV} features, improving computational efficiency while preserving spatial context.
    
    \item \textbf{Spatial-Temporal Attention}: Beyond spatial context, temporal consistency is essential for reasoning about environmental dynamics~\cite{bharilya2024machine}. The temporal encoder processes sequential \gls{BEV} features, employing mechanisms such as cross-attention~\cite{vaswani2017attention} to model temporal relationships between frames. This yields spatial-temporal feature representations that capture motion patterns over time.
    
    \item \textbf{Map Encoder}: \gls{HD Map} and traffic light information play a crucial role in constraining and guiding motion prediction~\cite{ettinger2021large}. Map data are transformed into the ego-vehicle’s coordinate frame and cropped to a predefined spatial extent. Traffic light status is incorporated into the map representation as an environmental indicator. Both are encoded into neural feature embeddings that can interact with spatial-temporal features from the LiDAR stream.
    
    \item \textbf{Multi-Agent Interaction Attention}: Accurately modeling interactions among agents remains a core challenge in motion forecasting~\cite{bharilya2024machine}. This module fuses agent features with map features to capture agent–map relationships, followed by inter-agent attention within \gls{ROIs} regions to model social interactions that influence future trajectories.
    
    \item \textbf{\gls{PandP} Decoder}: The decoder aggregates spatial-temporal, map, and interaction features to jointly output detection and prediction results. Detection is represented as a \gls{BEV} map containing a static map mask and dynamic object masks. Prediction is represented as a \gls{BEV}-based flow field encoding the estimated future trajectories of agents, facilitating downstream modules such as planning and decision-making.
\end{itemize}

\begin{figure}[t]
\centering
\includegraphics[width=0.6\linewidth]{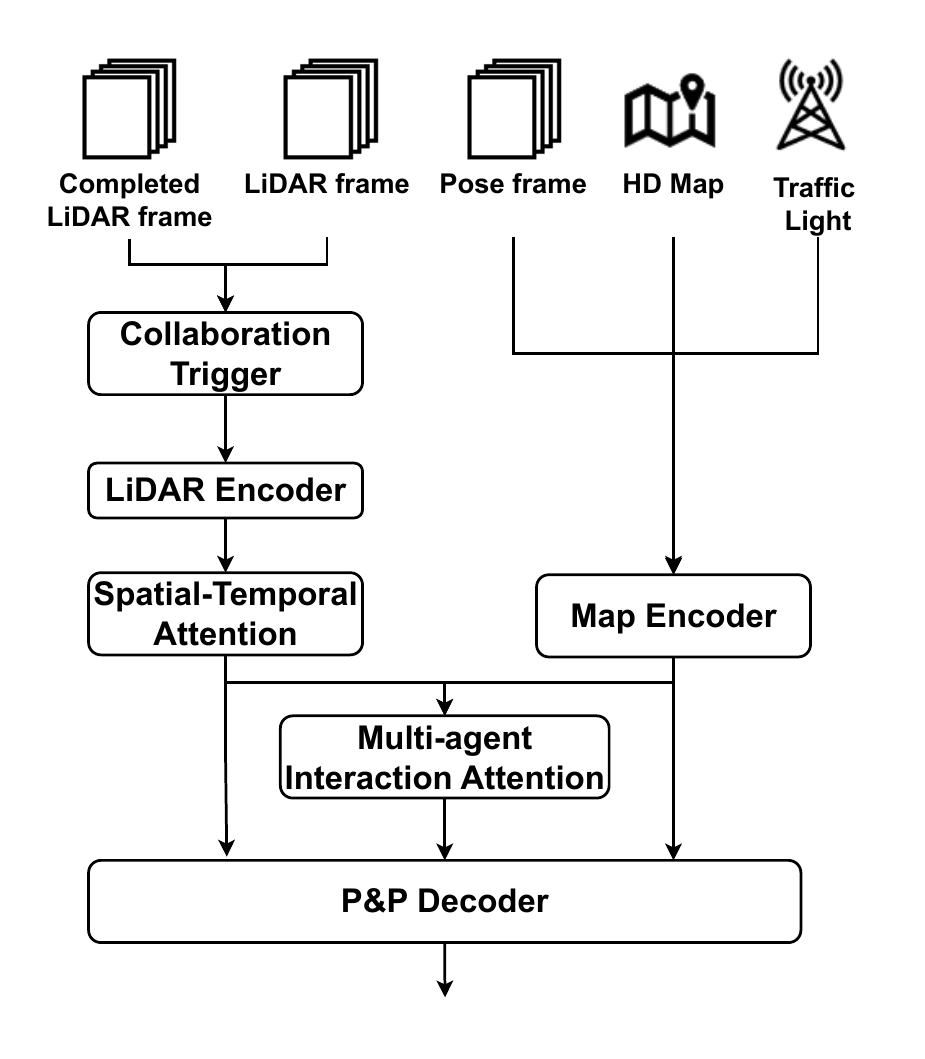}
\caption{Conceptual diagram of the \gls{PandP} module, showing the LiDAR encoder, spatial-temporal attention, map encoder, multi-agent interaction attention, and \gls{PandP} decoder.}
\label{fig:pandp_framework}
\end{figure}

\subsection{Collaboration Trigger}

While multi-agent collaboration can substantially enhance the perception capabilities of \gls{CAVs}, it also introduces significant computational and communication overhead. Collaboration may be unnecessary in situations where the ego vehicle has unobstructed visibility of relevant objects. To balance performance gains against resource consumption, a collaboration trigger is used to selectively activate collaborative modules only when beneficial. Designing such a trigger and identifying effective decision factors remain open research challenges~\cite{huang2023v2x}. In our framework, the trigger metric considers the degree of environmental occlusion along the driving path, the confidence level of the ego vehicle’s perception results, and the prevailing communication conditions. The collaboration module is activated once the computed metric exceeds a predefined threshold.

\subsection{Collaborative Scene Completion} \label{sec:co_scene_complete}

Conventional \gls{CP} frameworks typically exchange either neural features extracted by task-specific \gls{CP} models or the final perception results among agents. While effective for individual tasks, this approach lacks generality, the shared information can only support the task for which it was generated. This task dependence creates heterogeneity in the system and limits interoperability among diverse agents~\cite{han2023collaborative}. Furthermore, such methods often require joint model training across agents, and in some cases, re-training for each perception task~\cite{MultiRobotSceneCompletion--lib}, which can be costly and impractical in real-world deployments.

The proposed framework decouples \gls{V2X} collaboration from the \gls{PandP} pipeline, enabling each to be trained independently. We employ task-agnostic \gls{CSC} to generate a comprehensive, unified representation of the scene that can support multiple downstream tasks without requiring task-specific communication. This is achieved by reconstructing a complete scene from latent features and sharing only these features across agents, thereby reducing bandwidth usage. As shown in Figure~\ref{fig:csc}, each participating agent processes its original LiDAR frames through an encoder to generate compact latent feature representations. These features are exchanged between agents via \gls{V2X} communication and fused to enrich the scene representation. A decoder then reconstructs a completed LiDAR frame for each agent, filling in occluded or missing regions using information from collaborators. The resulting completed LiDAR frames provide a task-agnostic, dense environmental view that can be leveraged by multiple downstream perception and prediction tasks.

\begin{figure}[t]
\centering
\includegraphics[width=0.75\linewidth]{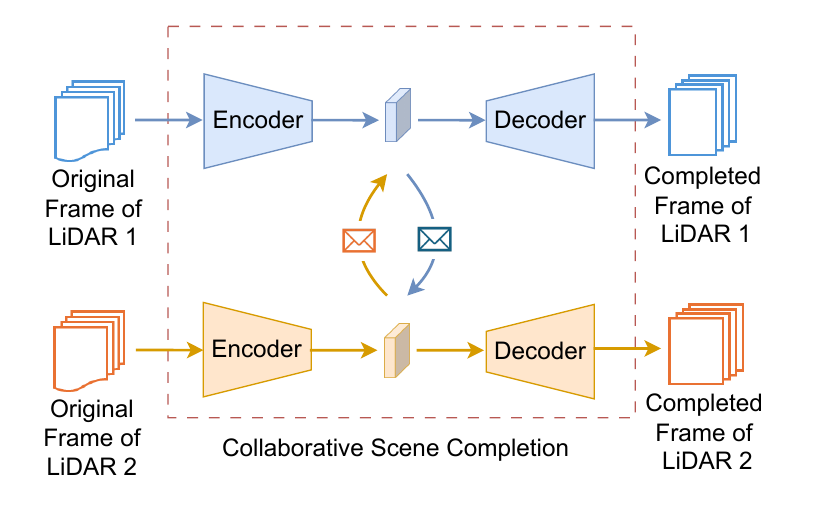}
\caption{Schematic of the CSC module: per-agent encoders produce latent features that are exchanged via V2X and decoded into completed LiDAR frames.}

\label{fig:csc}
\end{figure}

\section{Evaluation} \label{sec:evaluation}

In addition to developing the \gls{Co-PandP} system, it is crucial to establish effective evaluation methodologies that can accurately measure its performance. Evaluating \gls{CP} introduces unique challenges, as most existing studies adopt evaluation methods designed for single-vehicle perception. Such methods fail to capture one of the primary benefits of \gls{CP},its ability to mitigate visual occlusions~\cite{UMCUnifiedBandwidthefficient-2023-wangb}. To address this shortcoming, new evaluation strategies tailored for collaborative scenarios are required.

\begin{table}[t]
\centering
\caption{Summary of Metrics for Evaluating Joint Perception and Prediction \cite{vehits25}}
\label{tab:metrics}
\begin{tabular}{c|p{5cm}}
\toprule
Metrics & Description   \\ \midrule
$\mathrm{minADE}_k$ & the minimum over $k$ predictions of Average Distance Error: the average of point-wise L2 distances between the prediction and ground-truth forecasts   \\ \midrule
$\mathrm{minFDE}_k$ & the minimum over $k$ predictions of Final Distance Error: the L2 distance at the final future time step  \\ \midrule
$\mathrm{MR}_{k@x}$ & the MissRate: the ratio of forecasts having $\mathrm{minFDE}_k > x=4m$  \\ \midrule
$\mathrm{mAP}_f$ & the Mean Forecasting Average Precision: adapted from $\mathrm{mAP}_{det}$, $\mathrm{mAP}_f$ additionally penalizes trajectories that have correct first-frame detections but inaccurate forecasts ($\mathrm{minFDE}_k < 4m$), but also trajectories with incorrect first-frame detections (center distance $<2m$)   \\ 
\bottomrule
\end{tabular}
\end{table}

\subsection{Evaluation Method}

Since \gls{CP} primarily augments single-vehicle perception by addressing occlusion-induced blind spots, evaluation metrics should emphasize performance on objects that are invisible to the ego vehicle but visible from collaborating agents. For example, Wang et al.~\cite{UMCUnifiedBandwidthefficient-2023-wangb} propose the \textit{Average Recall of Collaborative View} (ARCV) metric, which measures recall for objects invisible from an ego-vehicle perspective yet detectable through collaboration. Following this principle, our evaluation categorizes objects into three visibility groups: \textit{fully visible}, \textit{partially visible}, and \textit{fully invisible}. We measure perception performance across these categories both without collaboration and with \gls{V2X} collaboration, quantifying the improvement achieved by \gls{CP}.

We deliberately report recall stratified by visibility rather than a visibility-stratified mAP, for a principled reason. Average precision combines recall with precision, and precision is governed by false positives, that is, predicted boxes that match no ground-truth object. A false positive has no associated ground-truth object and therefore cannot be assigned to a visibility group, since visibility is defined only on ground-truth objects. Partitioning precision, and hence mAP, by visibility is thus ill-posed, because the false-positive term cannot be coherently distributed across groups. Recall, by contrast, is computed entirely over ground-truth objects, each carrying a well-defined visibility label, which makes it the appropriate metric for isolating how collaboration recovers occluded objects. Designing a precision-aware metric that assigns unmatched detections to visibility groups in a principled way remains an open challenge.

In addition to perception accuracy, communication efficiency is a critical factor. We assess communication overhead using the \textit{average message size} metric, following \cite{marez2022bandwidth}, to capture the bandwidth requirements of collaborative methods.

Evaluating \gls{PandP} poses additional challenges, particularly when comparing traditional modular pipelines, where detection, tracking, and prediction are executed sequentially, with end-to-end frameworks that directly process raw sensor data to produce integrated perception and prediction outputs. To ensure fair comparison, both approaches are provided with identical detection and tracking inputs to the forecasting stage. Xu et al.~\cite{Xu2023TowardsMF} propose an evaluation strategy applicable to both paradigms, utilizing the metrics summarized in Table~\ref{tab:metrics}. A key measure is the \textit{mean forecasting average precision} ($\mathrm{mAP}_f$), inspired by object detection average precision ($\mathrm{mAP}_{det}$)~\cite{peri2022forecasting}.

\subsection{Evaluation with Synthetic and Real Dataset}

Benchmarking perception and prediction algorithms on datasets remains a fundamental practice for assessing and comparing methods. The datasets currently available for \gls{Co-PandP} are summarized in Table~\ref{tab:dataset}. For example, Wang et al. present DeepAccident, the first synthetic dataset designed to support \gls{Co-PandP}. It contains 57,000 cooperative frames from four simulated agents, generated using CARLA and SUMO, and includes annotations for various perception tasks such as detection, tracking, semantic segmentation, and motion prediction. While synthetic datasets offer scalability and annotation flexibility, they lack the environmental complexity and sensor noise present in real-world data. To address this, real-world datasets such as DAIR-V2X-Seq~\cite{V2XSeqLargeScaleSequential-2023-yub} provide valuable resources for \gls{Co-PandP} research. DAIR-V2X-Seq includes approximately 7,500 \gls{V2I} cooperative frames with paired infrastructure- and vehicle-side LiDAR point clouds and camera images.

However, \gls{PandP} models, particularly those based on deep learning, require large-scale datasets to achieve strong generalization. The limited size of current datasets constrains the training of larger and more complex models. To advance \gls{Co-PandP} research, there is a clear need for more extensive \gls{CP} datasets that provide broader scenario coverage and richer annotations.

We deliberately survey the complete landscape of currently available \gls{Co-PandP} datasets to make explicit the trade-offs that guide dataset selection in this emerging field. Among them, DAIR-V2X-Seq is the only public real-world dataset that jointly offers time-synchronized cooperative LiDAR, tracking and trajectory annotations, and HD-map
information; we therefore adopt it for our experiments, while synthetic alternatives (e.g., DeepAccident) and modality-limited datasets (e.g., LUCOOP) are discussed for completeness but are not suited to the full perception-and-prediction pipeline evaluated here.

\begin{table}[t]
\centering
\caption{Overview of all publicly available datasets for \gls{Co-PandP} that include infrastructure perspective. 
L indicates LiDAR and C denotes camera sensor in the modalities.}
\begin{tabular}{c|cccccc}
\toprule
Dataset      & Year & Collaboration & Modalities & Location & \# classes & \# co-frames \\ 
\midrule
DAIR-V2X-Seq \cite{V2XSeqLargeScaleSequential-2023-yub} & 2023 & V2I           & L \& C   & China    & 10         & 7.5k      \\
LUCOOP \cite{LUCOOPLeibnizUniversity-2023-axmanna}      & 2023 & V2V           & L        & Germany  & 4          & 13.5k     \\
DeepAccident \cite{DeepAccidentMotionAccident-2024-wanga} & 2024 & V2V, V2I & L \& C & CARLA    & 2          & 57k       \\
\bottomrule
\end{tabular}
\label{tab:dataset}
\end{table}

\section{Experimental Results} \label{sec:result}
\subsection{Experimental Setup}

\textbf{Dataset.} We conduct our experiments on DAIR-V2X-Seq~\cite{V2XSeqLargeScaleSequential-2023-yub} (also referred to as V2X-Seq), a real-world sequential dataset for vehicle--infrastructure cooperative perception and forecasting, built upon the DAIR-V2X benchmark and collected at urban intersections in the Beijing High-level Autonomous Driving Demonstration Area, China. We use its Sequential Perception subset, which provides more than 15{,}000 time-synchronized frames captured at 10\,Hz across 95 traffic scenarios (each lasting 10--20\,s), corresponding to approximately 7,500 cooperative vehicle--infrastructure frame pairs. Each cooperative frame contains a vehicle-side 40-beam LiDAR point cloud ($360^{\circ}$ horizontal FOV, $\leq 200$\,m range) and an infrastructure-side 300-beam roadside LiDAR point cloud ($100^{\circ}$ horizontal FOV, $\leq 280$\,m range, $\pm 3$\,cm accuracy), together with the corresponding $1920\times1080$ RGB images, calibrated sensor poses (vehicle GPS/IMU at 1000\,Hz), and 3D bounding-box annotations with consistent tracking IDs spanning 10 object categories. The dataset further provides vectorized HD map layers (lane centrelines and boundaries) which we feed to the map encoder of the prediction module. The point clouds are preprocessed by voxelization with a grid size of $[0.16, 0.16, 4]$\,m over the range $[-35\,\mathrm{m}, 35\,\mathrm{m}]$ along the $x$-axis and $[0\,\mathrm{m}, 70\,\mathrm{m}]$ along the $y$-axis.

\noindent\textbf{Implementation details.} We implement a modular perception and prediction pipeline, adopting PointPillars~\cite{lang2019pointpillars} for 3D object detection, AB3DMOT~\cite{weng2020ab3dmot} for multi-object tracking, and MTR~\cite{shi2022motion} for motion prediction. PointPillars is trained from scratch using the MMDetection3D \cite{mmdet3d2020} framework on the DAIR-V2X-Seq training split. We adopt the official PointPillars implementation and use the default training configuration provided by MMDetection3D, including the optimizer, learning rate schedule, batch size, and loss settings. MTR is trained independently using the official Valeo4Cast \cite{xu2024valeo4cast} implementation with its default training configuration. AB3DMOT is adopted from the original implementation and uses its default online tracking parameters without any additional training. All experiments are conducted on two NVIDIA L40S GPUs. Detection, tracking, and prediction models are trained once and reused across all three fusion strategies. Only the fusion stage differs between the compared settings; therefore, no retraining is performed for different fusion strategies, ensuring a fair comparison.

We implement multi-agent fusion at three points of the pipeline; in every
case the fused data are first transformed into a common reference
frame using the shared poses, which we refine by \emph{relative pose
optimization} (RPO). RPO corrects the residual misalignment between the two
agents that remains after GNSS-based localization. Inspired by the
registration step used in SLAM, we first associate objects co-observed by
the vehicle and the infrastructure (matching detection centers under the
initial relative pose) and then estimate the rigid-body relative-pose
correction, a rotation $\Delta R$ and a translation $\Delta\mathbf{t}$,
that minimizes the total distance between matched object centers:
\begin{equation}
(\Delta R^{\star}, \Delta\mathbf{t}^{\star})
= \operatorname*{arg\,min}_{\Delta R,\,\Delta\mathbf{t}}
\sum_{(i,j)\in\mathcal{M}}
\big\lVert \Delta R\,\mathbf{p}^{\text{veh}}_{i} + \Delta\mathbf{t}
- \mathbf{p}^{\text{inf}}_{j} \big\rVert_2^2 ,
\end{equation}
where $\mathcal{M}$ is the set of matched object pairs and
$\mathbf{p}^{\text{veh}}_{i}$, $\mathbf{p}^{\text{inf}}_{j}$ are the centers
of a matched object expressed in the vehicle and infrastructure frames,
respectively. This is the classical point-set registration objective used in
SLAM and can be solved iteratively in an ICP algorithm that alternates matching and minimization;
applying the resulting correction aligns the two agents' observations before
fusion, mitigating the localization errors that would otherwise corrupt it. Detection-level fusion operates on the per-agent 3D detections: the vehicle- and infrastructure-side bounding boxes are projected into the common frame and merged by non-maximum suppression at an IoU threshold of 0.5, keeping the highest-confidence box among overlapping detections. This injects complementary detections of occluded objects before any
temporal association, which is why it most improves recall on partially and fully occluded objects. Trajectory-level fusion operates after per-agent tracking: tracks from different agents are associated by Hungarian matching on center distance below 5m and combined with the track-to-track fusion rule of~\cite{625124} (covariance-weighted state fusion), exploiting temporal consistency while resolving cross-agent identity. Prediction-level fusion operates on the per-agent forecasts: the predicted trajectories of associated objects are aggregated across agents by confidence-weighted averaging, representing the latest possible fusion point. We include all three settings to isolate the effect of fusion timing: the detection and trajectory variants share information before the forecaster, whereas the prediction variant shares only final trajectories, which both increases bandwidth and forfeits the opportunity to repair upstream tracking errors.

\noindent\textbf{End-to-end prototype.} In addition to the modular pipeline, we implement a minimal \emph{end-to-end} Co-P\&P prototype that instantiates the two core operations of the framework, collaborative scene sharing and joint perception--prediction, within a single trainable model rather than a detection--tracking--prediction cascade. Collaborative sharing is realized with the RENO~\cite{you2025reno} neural point-cloud codec: the collaborating agent encodes its raw LiDAR sweep into a compact bitstream that is transmitted over \gls{V2X} and decoded on the ego side, where it is merged with the ego point cloud in the common reference frame using the RPO-refined poses described above. The fused point cloud is consumed by FutureDet~\cite{peri2022forecasting}, a jointly trained detection--forecasting model that outputs current detections together with their future trajectories in a single forward pass, thereby avoiding the inter-stage error accumulation of the modular pipeline. RENO exposes a single quantization parameter $q$ that trades reconstruction fidelity against bitrate; we sweep $q\in\{16,64,256\}$ (larger $q$ yields coarser quantization and higher compression) and additionally report the raw, uncompressed point cloud and an ego-only setting without communication. FutureDet is trained on the DAIR-V2X-Seq training split with its default configuration and forecasts the car category over the same horizon used for the modular pipeline. Notably, collaboration leads to some degradation in minFDE and MR, which may be attributed to RENO compression artifacts that have a limited impact on frame-level accuracy but are amplified during long-horizon trajectory extrapolation. This prototype is intentionally minimal: it realizes collaborative sharing through compression rather than through the learned, occlusion-completing CSC module, whose full validation we leave to future work.

\noindent\textbf{Evaluation metrics. } To evaluate detection performance, we use mean average precision (mAP) computed at an \gls{IoU} threshold of 0.5 and the recall at different visibility levels, thereby assessing the model’s ability to handle visual occlusions. Tracking performance is assessed using AMOTA metric, which extends the traditional MOTA metric by integrating it over multiple recall levels. It is computed as the average MOTA across a discrete set of recall thresholds, effectively summarizing false positives, false negatives, and identity switches across different confidence levels. Motion prediction is evaluated using the metrics summarized in Table~\ref{tab:metrics}, with $\mathrm{mAP}_f$ serving as the primary indicator, computed over a prediction horizon of $3$\,s ($30$ frames at $10$\,Hz). The communication cost is measured using the average message size in bytes, calculated by multiplying the number of transmitted objects by the size of each object message \cite{DAIRV2XLargeScaleDataset-2022-yua}.

\subsection{Quantitative Results}
To quantitatively assess the performance of different fusion strategies, we evaluate five settings: vehicle-only, infrastructure-only, and fusion applied at each of the three perception stages.

\begin{table}[t]
\centering
\caption{
Quantitative detection results on the DAIR-V2X-Seq dataset. 
Recall is reported for different visibility levels (Vis = fully visible, Part = partially visible, Invis = fully invisible). RPO = Relative Pose Optimization.
}
\label{tab:detection_results}
\begin{tabular}{l|ccc|ccc|ccc}
\toprule
\multirow{2}{*}{Method} 
& \multicolumn{3}{c|}{Recall$_{Car}$} & \multicolumn{3}{c|}{Recall$_{Cyclist}$} & \multicolumn{3}{c}{Recall$_{Ped}$} \\
& Vis & Part & Invis & Vis & Part & Invis & Vis & Part & Invis \\
\midrule
Vehicle only        & 67.7 & 64.9 & 52.3 & 53.5 & 60.6 & 41.5 & 49.4 & 41.9 & 35.5 \\
Infrastructure only & 75.3 & 45.0 & 57.2 & 45.0 & 30.5 & 11.7 & 14.8 & 12.1 & 17.5 \\
Late fusion w/o RPO & 62.5 & 76.5 & 93.2 & 72.0 & 71.6 & 85.0 & 66.3 & 70.6 & 61.3 \\
Late fusion w/ RPO  & 62.9 & 75.5 & 91.0 & 68.0 & 72.9 & 88.1 & 69.3 & 72.9 & 62.3 \\
\bottomrule
\end{tabular}
\end{table}

\begin{table}[t]
\centering
\caption{
Average Precision (AP) comparison of different methods on the DAIR-V2X-Seq dataset. 
AP is reported per class: IoU=0.5 for Car, IoU=0.3 for Cyclist and Pedestrian. All values are in percentage (\%).}
\label{tab:ap_results_all_methods}
\begin{tabular}{l|ccc}
\toprule
Method & Car AP@0.5 & Cyclist AP@0.3 & Pedestrian AP@0.3 \\
\midrule
Vehicle only         & 62.1 & 42.7 & 35.7 \\
Infrastructure only  & 63.5 & 40.7 & 16.1 \\
Late fusion w/o RPO  & 65.3 & 42.7 & 38.6    \\
Late fusion w/ RPO   & 65.9 & 51.6 & 39.6   \\
\bottomrule
\end{tabular}
\end{table}

As shown in Table~\ref{tab:ap_results_all_methods}, detection-level fusion improves performance over both vehicle-only and infrastructure-only baselines, achieving 65.3 AP for cars. With relative pose optimization (RPO), the AP further increases to 65.9. Additionally, Table~\ref{tab:detection_results} highlights significant improvements in recall for partially and fully occluded objects, indicating the effectiveness of late fusion in addressing visual occlusion.

\begin{table}[t]
\centering
\caption{Tracking performance for the \textbf{Car} category on the DAIR-V2X-Seq dataset.
Metrics include AMOTA, MOTP (both in percentage), and identity switches (IDS).}
\label{tab:car_tracking_metrics}
\begin{tabular}{l|ccc}
\toprule
Method & AMOTA ↑ & MOTP ↑ & IDS ↓ \\
\midrule
Vehicle only         & 58.3 & 32.4 & 26   \\
Infrastructure only  & 78.2 & 36.6 & 2526 \\
Track fusion w/o RPO & 72.5 & 39.8 & 12   \\
Track fusion w/ RPO  & 78.2 & 42.7 & 12   \\
\bottomrule
\end{tabular}
\end{table}

Table~\ref{tab:car_tracking_metrics} presents tracking results for the car class. Trajectory-level fusion (Track fusion) outperforms the vehicle-only baseline, reaching 72.5 AMOTA and significantly reducing identity switches. With pose optimization, AMOTA further improves to 78.2, matching the infrastructure-only baseline. These results demonstrate that tracking-level fusion effectively exploits temporal consistency and complementary viewpoints to maintain robust trajectories, especially in occluded scenes.

\begin{table}[t]
\centering
\caption{Motion prediction performance for the \textbf{Car} category on the DAIR-V2X-Seq dataset. 
Metrics include minimum average displacement error (minADE), minimum final displacement error (minFDE), miss rate (MR), and forecast mAP ($\mathrm{mAP}_f$).}
\label{tab:car_prediction_metrics}
\begin{tabular}{l|cccc}
\toprule
Method & minADE ↓ & minFDE ↓ & MR ↓ & $\mathrm{mAP}_f$ ↑ \\
\midrule
Vehicle only         & 18.9 & 37.9 & 1.0 & 4.9 \\
Infrastructure only  & 16.3 & 29.4 & 0.9 & 9.8 \\
Late fusion w/o RPO  & 14.3 & 26.3 & 0.8 & 10.2 \\
Late fusion w/ RPO   & 14.8 & 30.3 & 0.8 & 9.7 \\
\bottomrule
\end{tabular}
\end{table}

Table~\ref{tab:car_prediction_metrics} summarizes motion prediction results. Unlike detection and tracking, prediction-level fusion does not yield substantial improvements. This is largely due to the sensitivity of motion prediction models to upstream tracking errors such as identity switches, which fragment trajectories and hinder long-term forecasting accuracy.

\begin{table}[t]
\centering
\caption{Comparison of different fusion stages on overall performance for the \textbf{Car} category on the DAIR-V2X-Seq dataset. 
Detection results are reported as mAP (Car@0.5), tracking results as AMOTA, and prediction results as $\mathrm{mAP}_f$. 
RPO = Relative Pose Optimization. 
Det. fusion = Detection-level fusion, Track. fusion = Trajectory-level fusion, Pred. fusion = Prediction-level fusion. A ``/'' indicates a metric that is not affected by fusion at that stage, because the fusion is applied downstream of where the metric is measured (e.g., trajectory-level fusion leaves the detection AP unchanged).}
\label{tab:fusion_stage_comparison}
\begin{tabular}{lcccc}
\toprule
Fusion Stage & AP@0.5 ↑ & AMOTA ↑ & $\mathrm{mAP}_f$ ↑ & Comm. Cost (Bytes) ↓ \\
\midrule
Vehicle only  & 62.1 & 58.3 & 4.9 & 0 \\
Infrastructure only & 63.5 & 78.2 & 9.8 & 0 \\
Det. fusion   & 65.3 & 73.7 & 10.3 & 346.3 \\
Track. fusion & / & 78.2 & 12.2 & 406.3 \\
Pred. fusion  & / & / & 9.7 & 4003.9 \\
\bottomrule
\end{tabular}
\end{table}

Finally, Table~\ref{tab:fusion_stage_comparison} summarizes the trade-offs between performance and communication cost across different fusion stages. Detection-level fusion improves detection AP over both the vehicle-only and infrastructure-only baselines, and tracking-level fusion improves AMOTA over the vehicle-only baseline, while both introduce only moderate communication overhead. Gains over the stronger infrastructure-only baseline are smaller and, for some metrics, fall within the expected run-to-run variation, so we do not claim uniform superiority across all metrics. In contrast, prediction-level fusion requires significantly higher bandwidth (4003.9 bytes per message) without yielding corresponding performance gains. Because DAIR-V2X-Seq is sampled at 10\,Hz, these per-frame payloads correspond to sustained rates of roughly $3.5$\,KB/s (detection), $4.1$\,KB/s (tracking) and $40.0$\,KB/s (prediction), making the bandwidth penalty of prediction-level fusion explicit in real-time terms. These findings suggest that, for practical deployment, fusion should be performed prior to motion prediction, ideally at the detection or tracking level, where the balance between accuracy and communication cost is most favorable.

\begin{table}[t]
\centering
\caption{End-to-end Co-P\&P prototype on the DAIR-V2X-Seq \textbf{Car} category: collaborative point-cloud sharing via the RENO neural codec feeding a jointly trained FutureDet detection--forecasting model. \texttt{RENO\,$q$} denotes the codec quantization level (larger $q$ = stronger compression). Bytes/frame is the per-frame payload transmitted by the collaborating agent and Comp.\ the compression ratio relative to the raw point cloud. $\mathrm{mAP}_f$ is reported in percentage. Best value per metric in \textbf{bold}.}
\label{tab:e2e}
\begin{tabular}{l|cc|cccc}
\toprule
Condition & Bytes/frame ↓ & Comp.\ ↑ & $\mathrm{mAP}_f$ ↑ & minADE ↓ & minFDE ↓ & MR ↓ \\
\midrule
Ego (no comm.)     & \textbf{0}        & --                    & 14.15          & 3.561          & 6.216          & \textbf{0.448} \\
RENO $q256$        & 19{,}898          & \textbf{34.3$\times$} & \textbf{14.94} & 3.539          & 6.242          & 0.452 \\
RENO $q64$         & 63{,}151          & 10.8$\times$          & 14.88          & \textbf{3.527} & 6.222          & 0.450 \\
RENO $q16$         & 106{,}221         & 6.4$\times$           & 14.80          & 3.529          & \textbf{6.221} & 0.451 \\
Raw (uncompressed) & 682{,}712         & 1.0$\times$           & 14.79          & 3.530          & 6.222          & 0.451 \\
\bottomrule
\end{tabular}
\end{table}

\noindent\textbf{End-to-end prototype.} Table~\ref{tab:e2e} reports the end-to-end prototype on the car category, and two observations stand out. First, \emph{collaboration helps}: enabling point-cloud sharing raises forecasting accuracy from $\mathrm{mAP}_f=14.15$ (ego, no communication) to $14.94$ and lowers minADE from $3.561$ to $3.527$--$3.539$, confirming that complementary infrastructure observations benefit a jointly trained forecaster, not only the modular pipeline. Second, \emph{the shared representation can be made bandwidth-efficient}: the strongest setting (RENO $q256$) compresses the transmitted point cloud by $34.3\times$ (from $682{,}712$ to $19{,}898$ bytes per frame) while \emph{improving} $\mathrm{mAP}_f$ over both the raw transmission and the ego-only baseline, and accuracy stays essentially flat across the quantization sweep (within $0.15$ $\mathrm{mAP}_f$ points), indicating that aggressive neural compression preserves the task-relevant geometry. Crucially, the absolute forecasting accuracy of this jointly trained prototype (minADE $\approx 3.5$, minFDE $\approx 6.2$) far exceeds that of the modular pipeline (minADE $14$--$19$, minFDE $26$--$38$ in Table~\ref{tab:car_prediction_metrics}), even though both are trained and evaluated on the same data. This isolates cascade error accumulation, rather than the dataset or the forecaster itself, as the dominant cause of the modular pipeline's weak forecasting and of the futility of prediction-level fusion.

\section{Discussion} \label{sec:discussion}

Our experiments consistently show that fusing collaborative information
before the prediction stage is preferable to fusing predicted
trajectories. The underlying reason is that prediction-level fusion
operates on the output of an already error-prone tracking stage:
identity switches and trajectory fragmentation cannot be repaired by
aggregating final forecasts, and they propagate directly into the
displacement and $\mathrm{mAP}_f$ metrics. Detection- and tracking-level fusion,
in contrast, enrich the scene representation while temporal association
is still being formed, allowing complementary viewpoints to suppress
occlusion-induced misses before they corrupt the forecast. This
interpretation is reinforced by the communication analysis:
prediction-level fusion not only fails to improve accuracy but also
incurs the highest bandwidth, making it the least favourable option on
both axes.
 
\emph{Why earlier fusion outperforms prediction-level fusion.} Our finding is consistent with the broader collaborative-perception
literature, in which intermediate/feature-level fusion (e.g.,
V2VNet~\cite{V2VNetVehicletoVehicleCommunication-2020-wangb}, V2XPnP~\cite{zhou2024v2xpnp}, UniV2X~\cite{yu2025end}) is
generally favoured over late fusion of high-level outputs. Whereas those
works learn fusion end-to-end, our study isolates the \emph{stage} of
fusion within a modular pipeline and shows that the same qualitative
ordering holds, providing a controlled, interpretable baseline that
complements end-to-end results.
 
For deployable Co-P\&P systems, these results argue for placing
collaboration at the detection or tracking level, and for activating it
selectively via the collaboration trigger so that the bandwidth cost of
sharing is incurred only when occlusion or low perception confidence
warrants it. This directly motivates the decoupled CSC\,+\,P\&P design,
in which a task-agnostic shared representation is produced upstream
rather than exchanging task-specific prediction outputs.
 
The experiments in this paper instantiate and benchmark the modular form
of the pipeline and the fusion-timing question; the CSC module and the
fully end-to-end P\&P architecture are conceptual contributions whose
empirical validation is left to future work, contingent on a
sufficiently large cooperative dataset. We further note that the absolute
prediction accuracy is limited by tracker fragmentation and the long
forecasting horizon, and that performance on vulnerable road users
(pedestrians, cyclists) lags that on vehicles. Finally, all experiments
are conducted on data recorded in China; because the evaluated methods
are geometry-based and rely on the HD map rather than region-specific
behavioural priors, the relative ordering of fusion stages is expected to
transfer, whereas absolute cross-region generalization remains to be
verified on multi-region data.

\emph{End-to-end versus modular.}
This account is corroborated by our end-to-end prototype (Table~\ref{tab:e2e}):
the jointly trained detection--forecasting model attains far lower displacement
errors than the modular cascade on the same data (Section~\ref{sec:result}),
isolating cascade error accumulation, particularly tracker fragmentation, rather
than the dataset or the forecaster, as the primary cause of the modular
pipeline's weak forecasting and of the futility of prediction-level fusion. The
prototype additionally shows that the shared scene representation can be
compressed by an order of magnitude with no loss of forecasting quality,
underscoring the bandwidth efficiency of the upstream, task-agnostic sharing
advocated above.
 
\section{Challenges} \label{sec:challenges}

While \gls{Co-PandP} holds significant potential for enhancing vehicle awareness in complex traffic scenarios, by mitigating accumulated perception errors and alleviating occlusion-related issues, its deployment in real-world environments remains challenging. The key challenges are outlined below.

\begin{itemize} 
\item \textbf{Localization errors}: Accurate sensor data fusion requires aligning all observations within a shared coordinate frame, which depends on precise vehicle localization. However, \gls{GNSS}-based localization typically exhibits positional errors ranging from 1~m to 3~m, potentially causing substantial misalignments that degrade fusion performance. Addressing these pose estimation errors is crucial for achieving reliable collaborative scene completion in our framework. Our experiments make this concrete: relative pose optimization (RPO) measurably changes fusion outcomes (Tables~\ref{tab:ap_results_all_methods}), confirming
that residual pose error directly affects collaborative perception in our
framework.

\item \textbf{Asynchrony}: Collaborative scene completion is complicated by temporal misalignment between observations from multiple agents. To reconstruct a coherent scene, information from other viewpoints is often necessary, yet these inputs may arrive asynchronously relative to the ego vehicle’s sensing timeline. Such temporal offsets can lead to inconsistencies in the perceived positions of dynamic objects. Robust mechanisms for handling asynchronous multi-agent data are therefore critical for accurate scene reconstruction.

\item \textbf{Domain shift}: In practical deployments, vehicles may use heterogeneous LiDAR sensors, such as spinning and oscillating LiDARs. Differences in scan patterns produce distinct point cloud characteristics, introducing domain shifts that can disrupt downstream perception pipelines~\cite{DIV2XLearningDomainInvariant-2023-xiangb,HPLViTUnifiedPerception-2023-liuc}. Developing scene completion methods that can adapt to each sensor’s unique data representation is essential for maintaining robust performance across platforms.

\item \textbf{Dependency on large-scale labeled datasets}: The \gls{PandP} module in the framework relies on a unified neural network architecture, avoiding hand-crafted processing stages. While this design simplifies the pipeline, it also increases the demand for large, high-quality labeled datasets during training, similar to other end-to-end driving models. Reducing reliance on manually annotated data remains a key research priority to facilitate scalable and cost-effective \gls{PandP} deployment. Consistent with our results, the limited scale of DAIR-V2X-Seq is a likely cause of the modest absolute forecasting accuracy we observe; we cannot yet verify this hypothesis without a larger dataset.
\end{itemize}

\section{Conclusion}
\label{sec:conclusion}

This paper presents a conceptual framework for \gls{Co-PandP}, consisting of two core components: collaborative scene completion and a \gls{PandP} module. By decoupling \gls{V2X} collaboration from perception, the framework enables independent training and validation of each module, facilitating scalable deployment in real-world applications. 

Our comparative study of different fusion strategies shows that fusing information before the motion prediction module enhances final prediction performance more effectively than fusing motion prediction results. This finding provides valuable insight into optimal fusion timing within the perception pipeline. To complement the modular study, we additionally implement and benchmark a minimal end-to-end Co-P\&P prototype that realizes collaborative point-cloud sharing with the RENO neural codec and joint detection--forecasting with FutureDet. This prototype confirms that collaboration improves forecasting accuracy and, importantly, that the shared scene representation can be compressed by roughly $34\times$ with negligible loss in accuracy, demonstrating that V2X-based collaborative perception and prediction can be both effective and bandwidth-efficient. Furthermore, by revisiting evaluation methodologies in \gls{CP}, we highlight that performance assessment across different object visibility levels is crucial, particularly for objects fully invisible to the ego vehicle but detectable via collaboration. Such metrics directly reflect \gls{CP}’s ability to overcome occlusions and should serve as a key motivation for adopting collaborative approaches.

The modular pipeline still exhibits notable shortcomings when handling small-scale objects such as pedestrians and cyclists, which should be addressed in future work. Particular emphasis should be placed on improving performance for these vulnerable road users. Beyond the algorithmic enhancement of individual components, future efforts will focus on constructing a large-scale dataset to advance \gls{Co-PandP} research. We have already recorded data from a test field, which will be annotated and released as a public dataset supporting \gls{Co-PandP} development. Additionally, there is an urgent need to design a new visibility-aware mAP evaluation metric, as current visibility-aware recall provides only partial insight.

\backmatter

% \bmhead{Supplementary information}
% If your article has accompanying supplementary file/s please state so here. 

\bmhead{Acknowledgements}

This work was supported by the German Federal Ministry for Economic Affairs and Climate Action (BMWK) within the program ``Novel Vehicle and System Technologies'' and the project ``Valid Innovative Comprehensive Sensor System for Cooperative Automated Driving'' (VALISENS), funding code 19A22009E.

\begin{appendices}

\section{Declarations}\label{secA1}

\begin{itemize} 
\item \textbf{Competing Interests}: Not Applicable
\item \textbf{Funding Information}: This work was supported by the German Federal Ministry for Economic Affairs and Climate Action (BMWK) within the program "Novel Vehicle and System Technologies" and the project "Valid Innovative Comprehensive Sensor System for Cooperative Automated Driving" (VALISENS), funding code 19A22009E.
\item \textbf{Author contribution}: All authors contributed to the study conception and design. Material preparation, data collection and analysis were performed by Lei Wan. The first draft of the manuscript was written by Lei Wan and all authors commented on previous versions of the manuscript. All authors read and approved the final manuscript.
\item \textbf{Data Availability Statement}: Not Applicable
\item \textbf{Research Involving Human and /or Animals}: Not Applicable
\item \textbf{Informed Consent}: Not Applicable
\end{itemize}

\end{appendices}

%%===========================================================================================%%
%% If you are submitting to one of the Nature Portfolio journals, using the eJP submission   %%
%% system, please include the references within the manuscript file itself. You may do this  %%
%% by copying the reference list from your .bbl file, paste it into the main manuscript .tex %%
%% file, and delete the associated \verb+\bibliography+ commands.                            %%
%%===========================================================================================%%
% \bibliographystyle{sn-mathphys-num}
\bibliography{main}% common bib file
\end{document}